# CBRW: A Novel Approach for Cancelable Biometric Template Generation based on 1-D Random Walk

*Nitin Kumar, Manisha*
*Department of Computer Science and Engineering,*
National Institute of Technology, Uttarakhand,
Srinagar Garhwal, Uttarakhand, India
manisharawatphd@nituk.ac.in, nitin@nituk.ac.in

*Abstract-* **Cancelable Biometric is a challenging research field in which security of an original biometric image is ensured by transforming the original biometric into another irreversible domain. Several approaches have been suggested in literature for generating cancelable biometric templates. In this paper, two novel and simple cancelable biometric template generation methods based on Random Walk (CBRW) have been proposed. By employing random walk and other steps given in the proposed two algorithms viz. CBRW-BitXOR and CBRW-BitCMP, the original biometric is transformed into a cancellable template. The performance of the proposed methods is compared with other state-of-the-art methods. Experiments have been performed on eight publicly available gray and color datasets i.e. CP (ear) (gray and color), UTIRIS (iris) (gray and color), ORL (face) (gray), IIT Delhi (iris) (gray and color), and AR (face) (color). Performance of the generated templates is measured in terms of Correlation Coefficient (Cr), Root Mean Square Error (RMSE), Peak Signal to Noise Ratio (PSNR), Structural Similarity (SSIM), Mean Absolute Error (MAE), Number of Pixel Change Rate (NPCR), and Unified Average Changing Intensity (UACI). By experimental results, it has been proved that proposed methods are superior than other state-of-the-art methods in qualitative as well as quantitative analysis. Furthermore, CBRW performs better on both gray as well as color images.**

1. **INTRODUCTION**

Biometric recognition [1] has been widely used for user authentication. However, there are privacy and security concerns associated with biometric recognition. One of the ways to enhance the privacy and security of traditional biometrics is provided by cancelable biometric. Different methods of cancelable biometric template generation proposed by researchers are discussed in research work [2]. In traditional biometric based application, user provides his biometric credentials to a sensor device. This sensor device will scan his biometric trait and discriminating features obtained from any machine learning method.

These extracted features are stored into some standalone database. This whole process is known as Enrolment process in traditional biometric based systems. Afterwards, when a user presents his biometric credential at sensor device, again same features are extracted same as enrolment process and matching is performed with the stored representations. This matching process results in either granting access to the user or denial of the access. This process is known as authentication. In contemporary scenarios with the evolution of new methods or techniques, it is possible to gain unauthorized access to the standalone database using software and/or hardware devices. It may result into compromise of the original biometric credentials. Due to the outspread of the corona virus disease (COVID), most activities are being performed in online mode which further increases the risk of private information such as user credentials i.e. username and password. Increased online activities have also increased the number of accounts corresponding to a single user i.e. accounts on e-commerce websites, internet banking and online learning platforms etc. and hence the need for user authentication has gone up sharply. If any of the sensitive user information such as password or biometric trait of one account is compromised, then it may also put on risk the security and privacy of other accounts. As humans have limited count of biometric traits such as iris, face, fingerprint, palmprint, ECG and voice etc. Hence, keep them secure from an adversarial attack is mandatory.

Cancelable biometric [2] plays an important role in providing security to biometric traits by transforming the original biometric in another irreversible domain. In cancelable biometric, instead of storing the original features directly into the database, firstly these features are destroyed or deformed by some transformation functions. The templates obtained after this transformation are totally meaningless in comparison to the original biometric as shown in Fig. 1. Further, these distorted versions of original biometric are stored in the database. There are four important characteristics [3] associated with these cancelable templates: (i) Diverse (ii) Non-invertible (iii) Revocable (iv) Performance. ***Diverse*** generally means each cancelable biometric template should be different from others or in other words no two persons should be allotted same biometric template. ***Non-invertible*** means original biometric features cannot be extracted with these cancelable biometric templates while ***Revocable*** property states that if cancelable biometric template of a person gets stolen or compromised then a new cancelable template will be allotted to him/her without compromising original biometric identity. ***Performance*** property states that cancelable biometric based system's recognition rate should be same as traditional biometric based systems. It simply means recognition rate should not deteriorate using cancelable biometric template. The word ***cancelable*** means if one's allotted cancelable biometric template gets compromised, a new template will be allotted to him by cancelling the previous one as happens in the case of Automatic Teller Machine (ATM) password. A strong transformation function is typically required for generation of strong cancelable biometric template.

      Random walk [4] (a.k.a. drunker's walk) is a method which has been extensively used in image segmentation [5,6]. In this method, a drunker randomly walks in left, right, up or down positions from the current position. Motivated by this concept, in this research

work, two cancelable biometric template generation methods based on random walk (CBRW) have been proposed. As a biometric image may be thought of as a matrix of intensity values, we have tried to exploit random walk model to transform the current intensity values into other intensity values such that the original biometric is distorted. A sample image of cancelable biometric template corresponding to a face image is depicted in

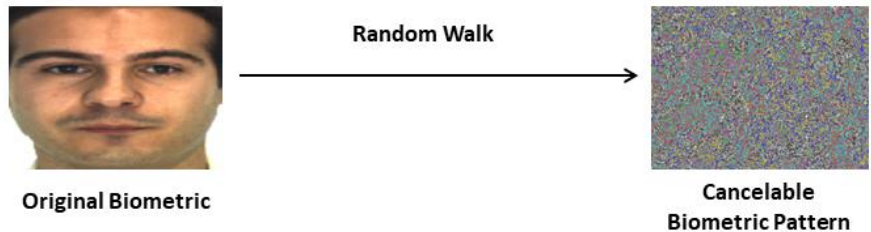

Fig. 1 Example of cancelable biometric template generated by proposed method

**1.1 Contribution**

In this paper, we have made following contributions:

i. Proposed two novel methods based on random walk model for generating cancelable biometric templates.
ii. Extensive experiments have been performed on eight publicly available gray and color biometric datasets.
iii. Proposed approaches are independent of the biometric trait used and hence experiments can be easily performed with other biometric modalities. In this work, experiments have been performed on three different biometric modalities i.e. face, iris and ear.

Besides the above contributions, there are some advantages of the proposed methods such as (a) Both methods are equally applicable to gray as well as color images (b) Performance of the proposed methods is superior than several other compared methods in qualative as well as quantitative terms. The rest of the paper is organized as follows: Section 2 briefly provides related work in cancelable biometric while Section 3 describes the proposed method in detail. In Section 4, experimental setup and results with qualitative, quantitative and histogram analysis are presented and discussed. Concluding remarks and future work are given in Section 5 at the end.

**2. Related Work**

In literature, various methods for cancelable biometric template generation have been suggested. These methods are further classified into six categories [2]: (i) Cryptography (ii) Transformation (iii) Filter (iv) Hybrid (v) Multimodal and (vi) Others. Each of these categories is further contained several template generation methods. Proposed methods comes under transformation based category which is one of the pioneer technique in the domain of cancelable biometric template generation in which original features are changed using different types of transformation such as Cartesian, Polar and Hadamard etc. In ***Cartesian transformation*** [3], image registration is required which is further followed by estimating position and orientation of singular points with respective minutiae positions. Fixed size cells are retrieved after dividing whole coordinate system. Cancelable biometric templates are formed by applying the changes in the cell positions. In Polar transformation [8], corresponding to the core position in polar coordinate, minutiae positions are calculated. In this transformation, polar regions are retrieved from the coordinate space. Further, changes are applied on these polar regions. Both these transformations are suffered from deviation problem such that if minutiae positions are changed in small scale in original biometric image, it can reflect a large deviation in the transformed minutiae domain. Fingerprint biometric is the most popular trait used in Cartesian and Polar transformation. ***Hadamard transform*** is a non-sinusoidal, orthogonal transformation which is used for projection on to walsh functions. Walsh functions are set of square or rectangular waveforms whose elements consists with values of +1 and -1, used for projection. Hadamard transformation used Hadamard matrix in which pairwise orthogonal row vectors have values +1 and -1. Two variants of Hadamard transform are known namely partial Hadamard and full Hadamard. Partial Hadamard [11] matrix is obtained after selecting some rows from full Hadamard and is non invertible in nature while full Hadamard transformation matrix is invertible in nature. Two important advantages of partial Hadamard includes low computational cost and low storage requirement. Uhl *et al.* [12] have generated cancelable iris templates which used wavelet transformation as a transformation key matrix. The main advantage of this method includes no occurrence of data loss and no requirement of features alignment. In ***Multiplicative transform*** [13], cancelable biometric templates are formed by element wise multiplication of original features with some random vector. Sorting method is applied on indexes of this resulting vector which is further used for retrieval of biometric template.

Pillai *et al.* [17] have used Sectored Random Projections for generation of cancelable iris template. In this approach iris image is divided into many parts which are known as sectors. Further, random projection is applied on individual sectors. Lastly, these transformed sectors are again concatenated to form cancelable iris template. Punithavathi *et al.* [16] have proposed extended version of this method in which individual sectors are projected on ***Dynamic Random Projection Matrix*** (DRPM) which finally generates cancelable iris template. Here, DRPM matrix is retrieved from the iris features which remove the requirement of any external matrix for transformation. Kim *et al.* [18] have used *Sparse Random Projection* for generation of cancelable face template. In this method, random matrix has the value -1, 0 and +1. The values used in random matrix is used for accelerating enrolment and authentication process.

## 3. Cancelable Biometric Template Generation based on Random Walk:

In this section, firstly generation of random walk matrix **R**$_W$ based on the popular random walk model is defined and then two methods for cancellable biometric template generation are proposed which are based on **R**$_W$. In the following, the original biometric image is denoted as S of size a×b and another image **R** which is generated using uniform distribution and is of the same size as the secret image.

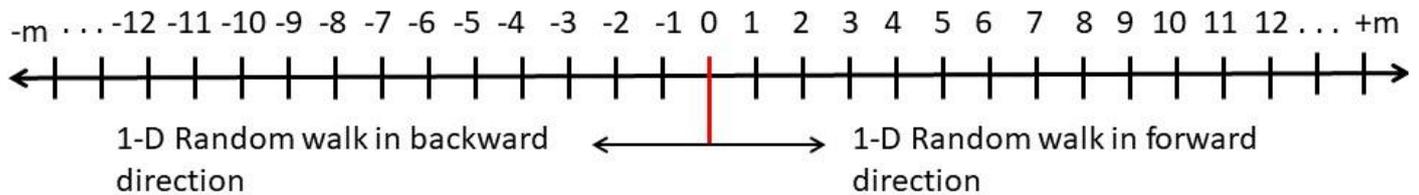

Fig. 2: Example of one dimensional horizontal walk

### 3.1 Generation of Random Walk Matrix **R**$_W$:

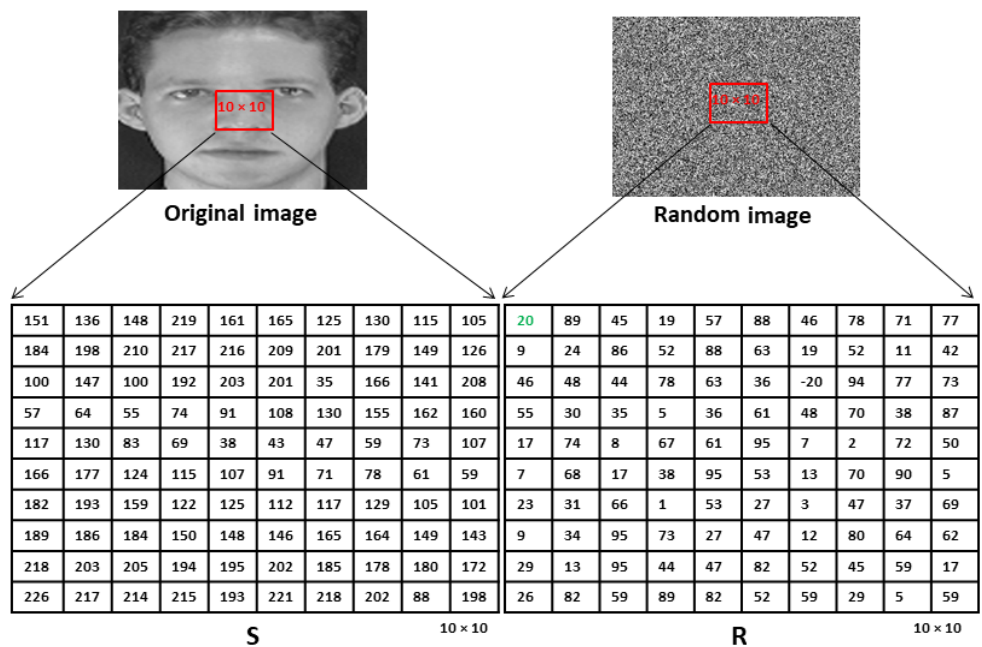

Fig. 3: Selection of 10×10 size sub-image from original biometric and random image and named them as secret image **S** and random matrix **R**

In random walk or drunker's walk method, a person is initially assumed to be present at the origin of the two dimensional plane and then he can move one step in any of the four directions i.e. left, right, up and down. Subsequently, he can move again in the four directions based on the position after the first step and so on. In this proposed work, one dimensional horizontal walk is used as shown in Fig 2. There are two directions for random movement i.e. (i) Forward (towards positive x-axis) (ii) Backward (towards negative x-axis). The main objective is to transform the pixel values in S with the help of random matrix **R** as shown for a sample sub-image of size 10 ×10 in Fig. 3. Based on random matrix **R**, the pixel values is modified in the original biometric image **S**. Suppose *p* and *r* are pixels location in matrix S and R respectively as shown in Fig. 3. Procedure for generation of $\mathbf{R}_W$ is given in Algorithm 1. In the proposed Algorithm, (i, j) is the location for any pixel in two dimensional entities i.e. where i and j represent row and column number respectively

**Algorithm 1**: Generation RWM(S, R)

   Input: Secret image **S** and random matrix **R**

   Output: Random walk matrix $\mathbf{R}_W$

   1 For i = 1 to a and j = 1 to b, repeat the following two steps

   2 If **R**(i, j) > 0 then **X**(i, j) = **S**(i, j) + **S**[(i, j) + **R**(i, j)] (See Fig. 4a).

   If **R**(i, j) < 0 then **X**(i, j) = **S**(i, j) + **S**[(i, j) - **R**(i, j)] (See Fig. 4c).

   3 $\mathbf{R}_W$(i, j) = **X**(i, j) mod 256

While moving forward, it may happen that we reach the lower rightmost pixel of **S** i.e. (a, b) and still we have to go forward. This condition is termed as Overflow and we start moving forward from the upper left corner of image (i.e. position (1,1) of the image) as illustrated in Fig. 4c. Similarly, while moving backward, it may happen that we reach the upper left corner i.e. location (1,1) in the image and we still need to move backward. In this case, the backward movement is done from the bottom rightmost pixel i.e. (a; b) backward. This condition is termed as Underflow. The Algorithm 1 above represents the eneration of random walk matrix for a gray scale image. For finding the random walk matrix for a color image, this Algorithm is applied on each of the channels (R,G,B). Next, two cancelable biometric template generation methods are discussed in which templates are generated with the help of $\mathbf{R}_W$ matrix. Step-by-step procedure of generation of random walk $\mathbf{R}_W$ matrix: For elaboration purpose 10×10 dimensions sub-image S is selected but experiments have been accomplished on image size of 320×240. The following steps will be applicable on any image of any dimensions for generation of $\mathbf{R}_W$ matrix:

1. Both **S** and **R** matrices should be of same dimensions. The **R** matrix contains both positive and negative integers. One null $\mathbf{R}_W$ matrix of same size as **S** and **R** is created for storing the transformed pixels. Pixels in **S** will move in forward and backward direction according to value in **R**.

2. $x$ is the pixel value at $p^{th}$ location in **S** and corresponding to same location, another pixel value $y$ is chosen from $r^{th}$ location in **R**.

3. If the pixel value $y$ at $r^{th}$ location in **R** is +ve then pixel value $x$ at $p^{th}$ location in **S** will move in forward direction. Similarly, for -ve pixel value in **R**, pixel value $x$ will step backward in **S**.

4. Let's take a small example of one pixel value $x$ at $p^{th}$ location in **S** has a value 151, corresponding to this pixel location, in **R** at $r^{th}$ location $y$ pixel value is 20 as shown in Fig. 4a. According to proposed work, $p^{th}$ pixel will move in forward direction 20 steps ahead and reaches at $q^{th}$ location which is at $21^{st}$ place and has pixel value $z = 100$. Now, add this pixel value $z = 100$ with original value $x$ which is 151, this will give $x + z = 251$. Further, to keep the pixel in between 0 to 255 range modulus operation by 256 is used. Finally, we got 251 as transformed pixel value of $x$ at $p^{th}$ location and stored in $\mathbf{R}_w$. The $x$ value at $p^{th}$ in original biometric has now transformed to value 251 instead of 151 in transformed domain and stored in corresponding location in $\mathbf{R}_w$. Similarly, all other pixels will also have converted in same manner.

5. In case of color image, each channel (R, G, B) pixel values are ranged between 0 and 255. For maintaining this pixel range we have used mod function. The effectiveness of mod operation is illustrated by following example. Suppose $x$ value in **S** is chosen at $4^{th}$ location and has pixel value 219, corresponding to this location, $y$ value in **R** is 19. As per proposed Algorithm 1, $x$ will move in forward direction 19 steps ahead in **S** and reached at location $23^{rd}$, which has pixel value $z = 100$. Further addition of these two values $x + z$ will result in 319. This resultant value is beyond standard pixel range. To resolve this problem, we have used mod function, 319 mod 256 results in 63. This 63 will now become new transformed value of $x$ at $4^{th}$ location in transformed domain and stored in $\mathbf{R}_w$.

6. A 10×10 dimensions' image results in total of 100 pixels. We need to perform transformation at $98^{th}$ location in **S**, which has pixel value 88 and corresponding to this location, **R** contains pixel value 5 at $98^{th}$ location. According to our method, we have to move 5 steps ahead in **S**, which results in pixel location $103^{rd}$, which is out of range as total number of pixels are 100. For handling this problem, we have provided solution of circular loop using mod function by 100 e.g. 103 mod 100, which results in 3. Now, in original biometric we stepped forward by 3 positions beyond $100^{th}$ location and reached at $3^{rd}$ location in **S**. The value at $3^{rd}$ place is 148. Now, addition is performed between 148 and 88, which results in new pixel value 236. Now 236 mod 255 is 236, which is to assign for $98^{th}$ location in transformed domain in R. The final value for $98^{th}$ pixel is 236 instead of 88 in transformed domain. Same solution will work for underflow condition with negative values in **R**, which leads to move in backward direction.

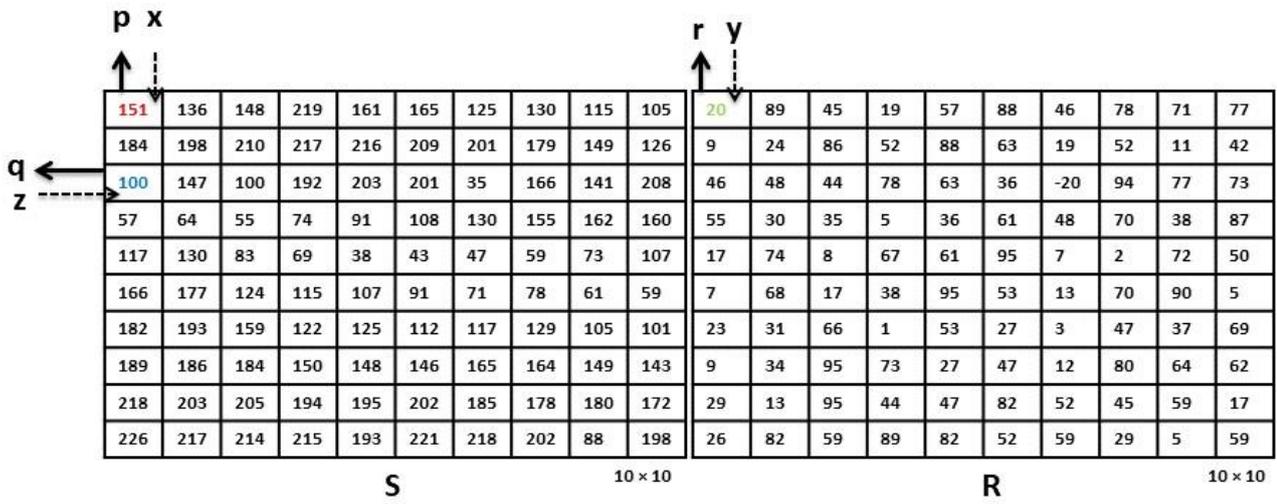
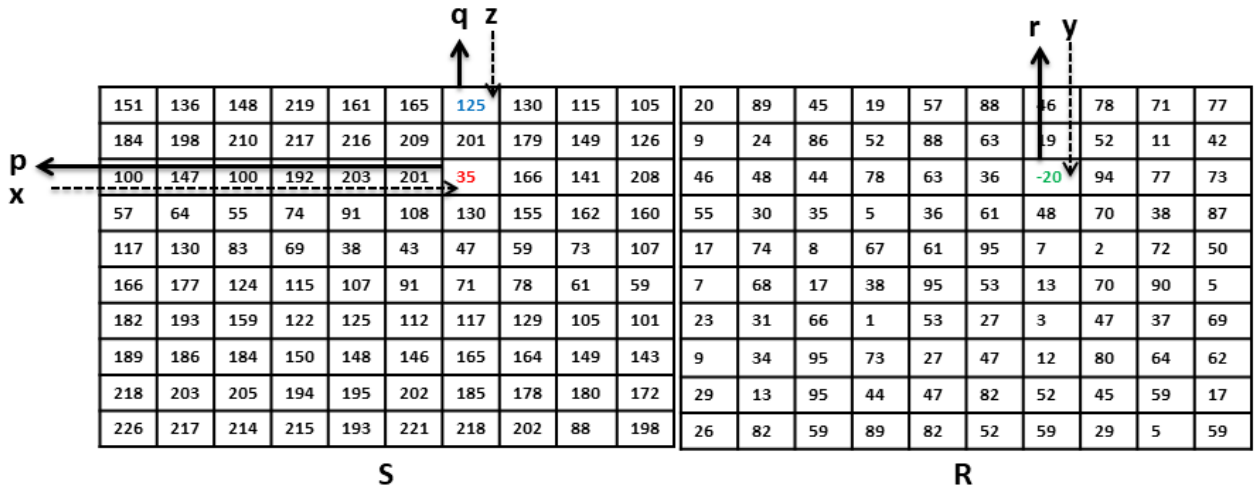
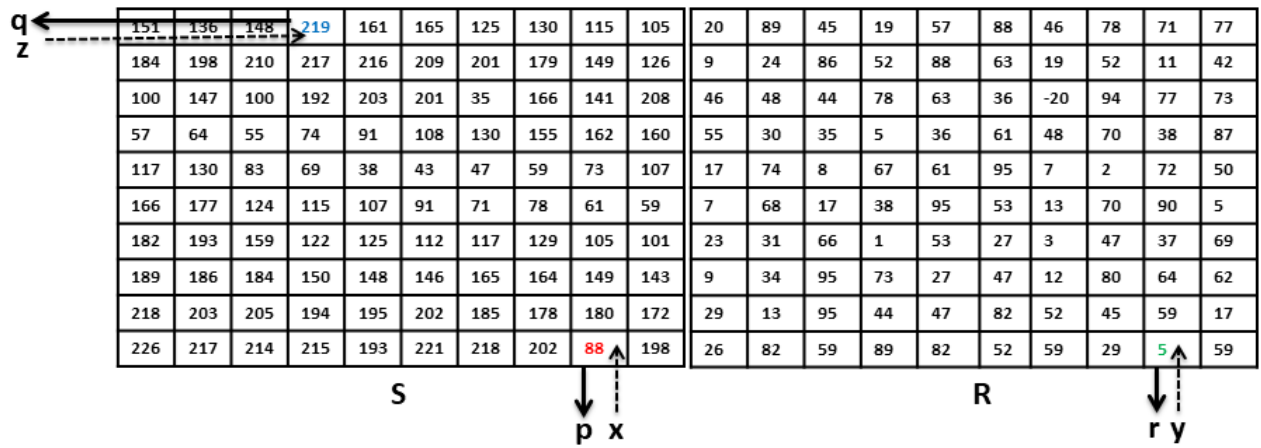

Fig. 4: Illustration for computation of $R_W$ matrix (a) $R(i,j)>0$ (b) $R(i,j)<0$ (c) *Overflow*

## 3.2 Cancelable Biometric Template based on Random Walk using Bit XOR (CBRW-BitXOR)

**Algorithm 2:** CBRW-BitXOR

Input: Original biometric image **S** and random image **R**

Output: Cancelable template **C**

1 $\mathbf{R}_W$ = Generate_RWM(**S**,**R**) // Obtain random walk matrix using Algorithm 1

2 **C**=**S**⊕$\mathbf{R}_W$ // Find the cancelable biometric template using XOR of **S** and $\mathbf{R}_W$

Here, we describe the method for generation of cancelable biometric template using BitXOR. The step-by-step procedure for cancelable biometric template generation using BitXOR method is given in Algorithm 2. Initially, the random walk matrix is obtained using Algorithm 1. Next, XOR operation between S and RW is performed for generation of cancellable biometric template. Some sample images of ear, iris and face biometrics together with their corresponding cancelable biometric templates generated by proposed methods (BitXOR and BitCMP) and other popular methods are shown in Fig 5. It can be seen from Figure that templates generated by proposed methods do not contain any traces of the original biometric while other methods do.

## 3.3 Cancelable Biometric Template based on Random Walk using Bit Complement (CBRW-BitCMP)

**Algorithm 3**: CBRW-BitCMP

Input: Original biometric image **S** and random image **R**

Output: Cancelable biometric template **C**

1 $\mathbf{R}_W$ = Generate_RWM(S,R) // Obtain random walk matrix using Algorithm 1

2 **I**=**S** ⊕$\mathbf{R}_W$ // Generate intermediate template

3 **C**=BitCMP(**I**) // Generate cancelable template

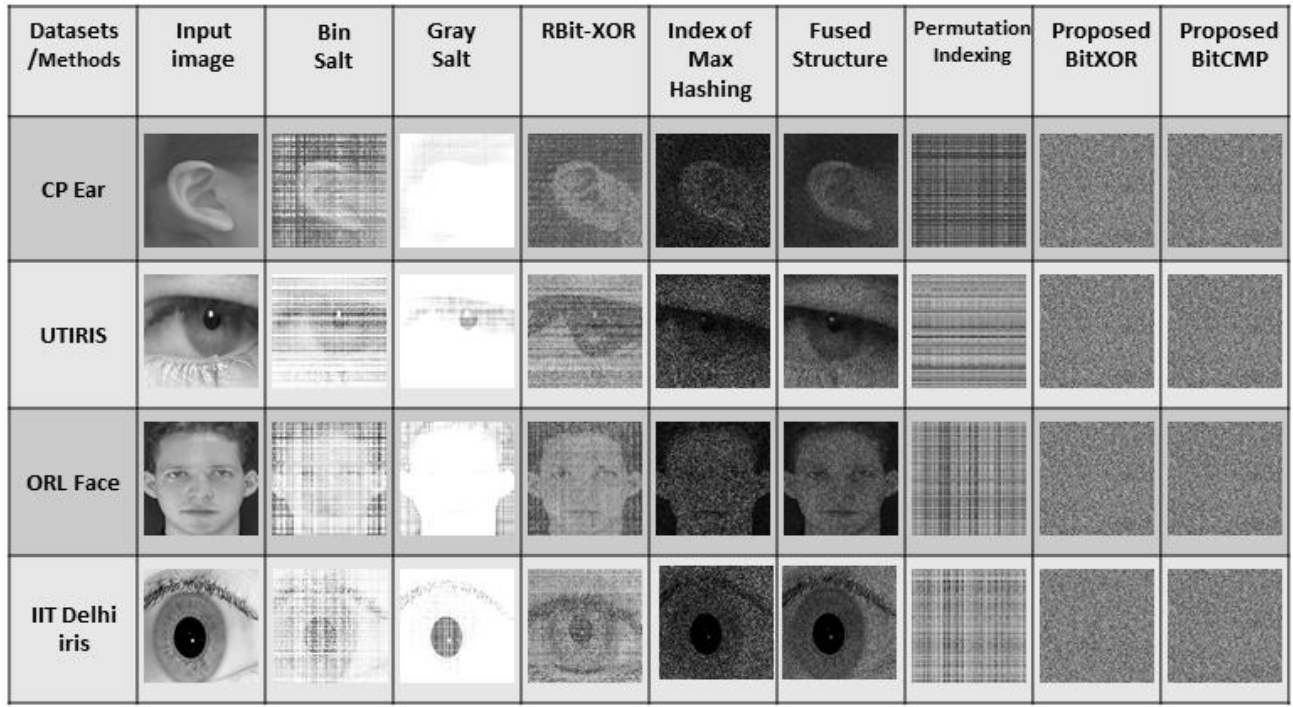

Fig. 5: Qualitative analysis among color cancelable biometric template generated using Bin Salt [27], Gray Salt [27], RBit-XOR [20], Index of Max Hashing [9], Fused Structure [10], Permutation Indexing [7], CBRW-BitXOR (proposed) and CBRW-BitCMP (proposed) on CP ear [25], UTIRIS [26], ORL face [22] and IIT Delhi iris [24] datasets

Table 1: Datasets Description

| Dataset | # Identities | Total Images | Image Size |
|---|---|---|---|
| ORL face [22] | 40 | 400 | 112 × 92 |
| AR face [23] | 126 | 4000 | 768 × 576 |
| IIT Delhi iris [24] | 224 | 1120 | 320 × 240 |
| CP ear (gray) [25] | 17 | 102 | 300×400 |
| CP ear (color) [25] | 60 | 185 | 80×150 |
| UTIRIS color [26] | 79 | 1540 | 2048 × 1360 |
| UTIRIS gray [26] | 79 | 1540 | 1000×776 |

Similar to the above Algorithm, in this subsection the method for generation of cancelable biometric template using Bit Complement is described. The proposed method uses random walk matrix $\mathbf{R}_W$ for generating cancelable biometric templates. Further, Algorithm 3 presents step-by-step procedure of template generation using BitXOR method. Initially, random walk matrix is obtained using Algorithm 1. Intermediate share $\mathbf{I}$ is generated by XOR operation performed between $\mathbf{S}$ and $\mathbf{R}_W$. Finally, the Bit Complement of the intermediate share is performed to get the cancelable biometric template. Some sample images of ear, iris, and face biometrics together with cancelable templates generated using proposed methods and other popular methods are shown in Fig 5. It can be seen from Figure that no original credentials are revealed by both proposed methods while other methods are unable to distort the original biometric without revealing them. The main difference between two proposed methods is that, the second method consist with one more step of bit complement BitCMP operation in addition to BitXOR operation whose output is taken as intermediate template I as described in Algorithm 3.

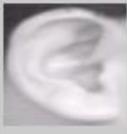

Fig. 6: Qualitative analysis among color cancelable biometric template generated using Bin Salt [27], Gray Salt [27], RBit-XOR [20], Index of Max Hashing [9], Fused Structure [10], Permutation Indexing [7], CBRW-BitXOR (proposed) and CBRW-BitCMP (proposed) on CP ear [25], UTIRIS [26], AR face [23] and IIT Delhi iris [24] datasets

## 4. Experimental Setup and Results

### 4.1 Dataset Description

Experiments have been performed of both proposed methods on three biometrics (both gray and color) viz. (i) Ear (ii) Iris and (iii) Face. The Carreira Perpinan(CP) [25] ear dataset is publicly available dataset which consists with total of 102 ears images of 17 peoples. This dataset consists of both gray and color images. The University of Tehran IRIS (UTIRIS) [26] dataset consists with 1540 iris images of 79 individuals. The dimensions of color dataset are 2048×1360 and for gray dataset dimensions are 1000×776. The Olivetti Research Laboratory (ORL) face [22] dataset formerly known as American Telephone & Telegraph company. This dataset is formed by 400 gray images of 40 different subjects with various facial expressions. Aleix Martinez and Robert Benavente (AR) face [23] dataset consists with 4000 images of 126 different individuals (70 males and 56 females). In IIT Delhi (IITD) iris [24] dataset (version 1.0) has 1120 iris images captured from 224 users (176 males and 48 females). A brief description about these datasets are provided in Table 1.

### 4.2 Qualitative Analysis:

Here, a qualitative analysis of cancelable biometric templates generated by proposed methods is presented. The quality of cancelable biometric templates generated by proposed method and other compared methods are shown in Fig. 5 for gray datasets while Fig. 6 shows cancelable biometric templates for color datasets. It can be seen from both these Figs. 5 and 6 that templates generated by proposed method for all datasets do not reveal any information about the original biometric while others methods do. A histogram analysis between original biometric **S** and cancelable biometric template **C** generated by proposed methods is also carried out. A histogram analysis between **S** and **C** is shown in Fig. 7 for gray datasets and Fig. 8 represents for color datasets using both proposed methods. It is easy to observe that histogram of cancelable template is attened in most cases (except the color iris image) in comparison to respective original image histogram. This analysis concludes that the intruder cannot gain any information from the cancelable biometric template as they are totally distorted.

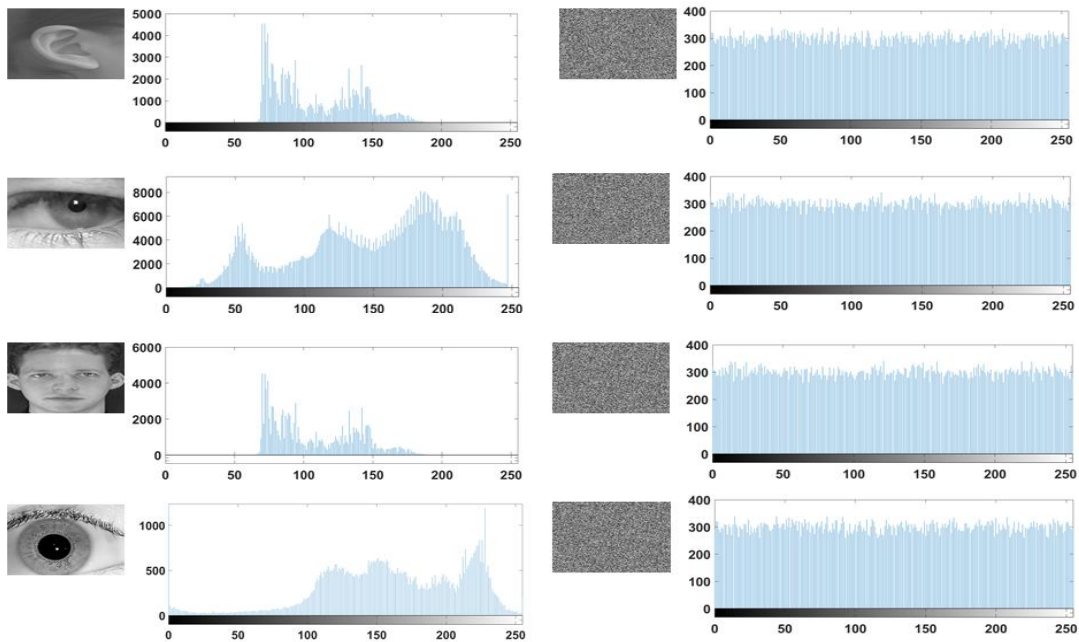

Fig. 7: Histogram analysis between gray original biometric S and cancelable biometric template C generated by CBRW-BitXOR (proposed) on CP ear [25], UTIRIS [26], ORL face [22] and IIT Delhi iris [24] datasets

## 4.3 Quantitative Analysis

A quantitative analysis is also carried out for cancelable templates generated by proposed methods and other compared methods. The performance of templates is measured in terms of Correlation (Cr), Mean Absolute Error (MAE), Number of Pixel Change Rate (NPCR), Peak Signal to Noise Ratio (PSNR), RMSE (Root Mean Square Error), SSIM (Structural Similarity) and UACI (Unified Average Changing Intensity).

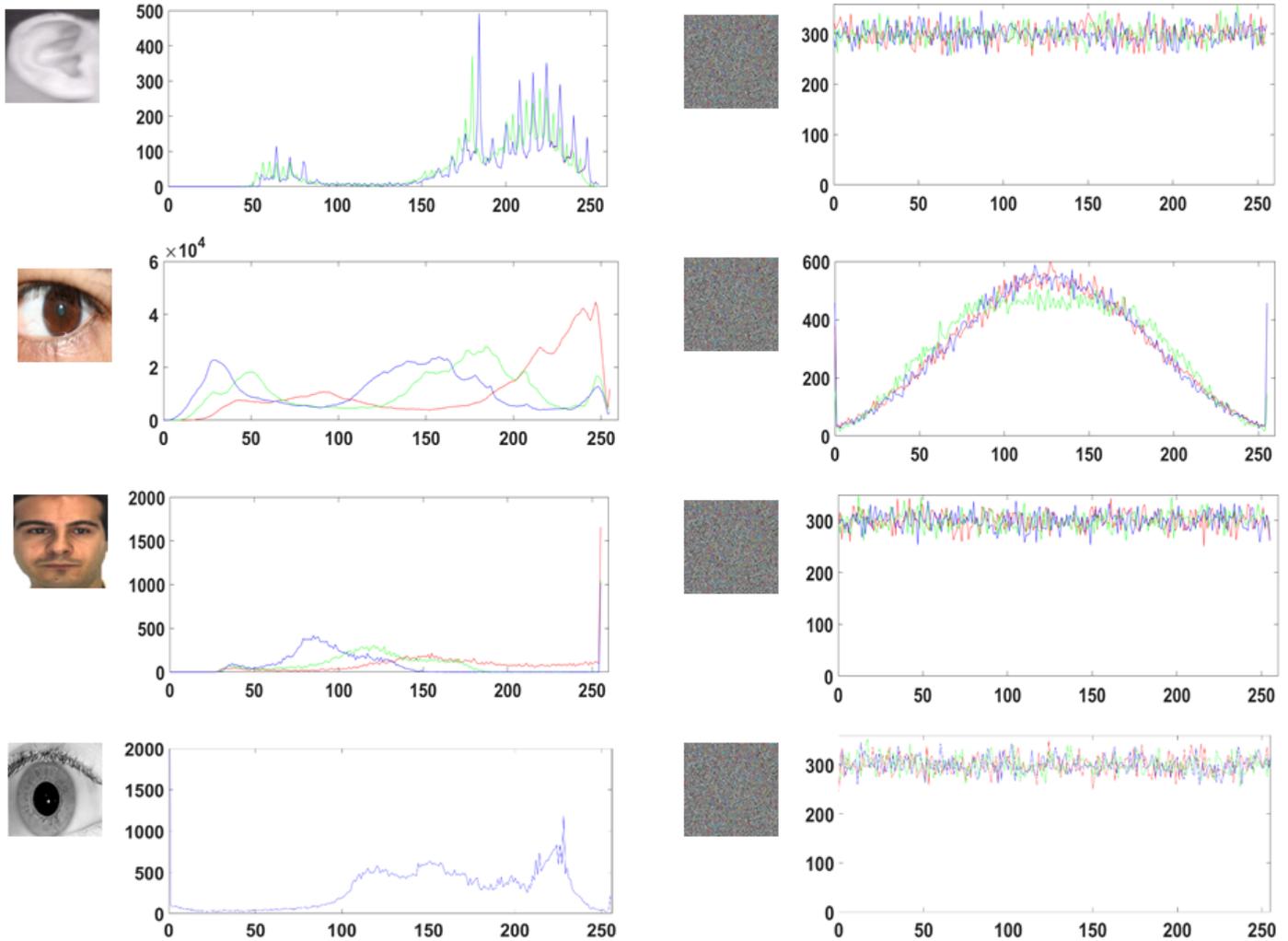

Fig. 8: Histogram analysis between color original Biometric **S** and color Cancelable Biometric **C** template by CBRW-BitXOR (proposed) on CP ear [25], UTIRIS [26], AR face [23], and IIT Delhi iris [24] datasets.

The ideal value for above mentioned measures for two similar images are given as: Correlation value is +1 and -1, MAE is 0, NPCR is 0, PSNR is infinite, RMSE is 0, SSIM is +1 and -1 and UACI is 0. In cancelable biometric based system, original biometric **S** and generated cancelable biometric template **C** should not be similar. Hence, the value for above mentioned performance measure in cancelable biometric domain should be opposite to ideal values. The values of these parameters in the case of cancelable biometric system should be minimum for Correlation, PSNR and SSIM. While for MAE, NPCR, RMSE, UACI values should be maximum. The computation formulae of above mentioned performance measures are given in Table 2. The quantitative results for both gray and color images are given in Tables 9-6. The detail description of seven performance measures is given below:

Table 2: Performance measures used in proposed work

| Measure | Formula | Description |
|---|---|---|
| Correlation Coefficient | $C_r = \frac{\sum_m \sum_n (S_{mn} - \bar{S})(C_{mn} - \bar{C})}{\sqrt{(\sum_m \sum_n (S_{mn} - \bar{S}))^2 - (\sum_m \sum_n (C_{mn} - \bar{C}))^2}}$ | Here S and C are original and cancelable template images respectively. |
| Mean Square Error | $\text{MSE} = \frac{1}{W \times H} \sum_{i=1}^{W} \sum_{i=1}^{H} (S(i,j) - C(i,j))^2$ | Here, $S(i,j) - C(i,j)$ is error and $W \times H$ image size. |
| Root Mean Square Error | $\text{RMSE} = \sqrt{MSE}$ | Here, $MSE$ is Mean Square Error |
| Peak Signal to Noise ratio | $PSNR(db) = 20 \log_{10} \frac{255}{\sqrt{MSE}}$ | Here, 255 is the maximum value of single channel. |
| Structural Similarity | $\text{SSIM}(S,C) = \frac{(2\mu_S \mu_C + T_1)(2\sigma_{SC} + T_2)}{(\mu_S^2 + \mu_C^2 + T_1)(\sigma_S^2 + \sigma_C^2 + T_2)}$ | $\mu_S$, $\mu_C$ are means and $\sigma_S$, $\sigma_C$ are variance and $\sigma_{SC}$ is covariance between images **S** and **C** respectively. And, $T_1$ and $T_2$ are constants |
| Mean Absolute Error | $\text{MAE} = \frac{1}{W \times H} \sum_{i=1}^{W} \sum_{i=1}^{H} |S(i,j) - C(i,j)|$ | Here, $S(i,j) - C(i,j)$ is error and $W \times H$ is image size or total number of pixels in an image |
| Number of Pixel Change Rate | $\text{NPCR}(S_i, C_j) = \frac{\sum_{i,j} D(i,j)}{W \times H} \times 100\%$ | Here $D(i,j) = 0$; if $S(i,j) = C(i,j)$ and $D(i,j) = 1$; if $S(i,j) \neq C(i,j)$ |
| Unified Average Changing Intensity | $\text{UACI} = \frac{1}{W \times H} \times \frac{\sum_{i=1}^{W} \sum_{i=1}^{H} |S(i,j) - C(i,j)|}{255} \times 100\%$ | Here, $S(i,j) - C(i,j)$ is error and $W \times H$ is image size or total number of pixels in an image. And, 255 is maximum color value in single channel. |

- **Correlation Coefficient**: It is used to present the relation between the two images. Two images are related to each other in terms of three types of correlation values i.e. (i) Zero (ii) Positive and (iii) Negative correlation. The possible value of correlation is between $-1.0$ to $+1.0$. The zero value of correlation between **C** and **S** denotes that they are totally dissimilar. However, the negative value of correlation denotes that the cancelable template **C** resembles with the negative of the original biometric image **S**. While, the positive value denotes that cancelable biometric template **C** resembles the original biometric **S**. In cancelable biometric system, correlation coefficient should be near to zero.

$$C_r = \frac{\sum_m \sum_n (S_{mn} - \bar{S})(C_{mn} - \bar{C})}{\sqrt{(\sum_m \sum_n (I_{mn} - \bar{I}))^2 - (\sum_m \sum_n (S_{mn} - \bar{S}))^2}} \quad (2)$$

green and blue) followed by taking their average i.e. divided by number of channels i.e. three.

- **Mean Square Error**: It denotes the sum of square of difference between the pixel intensity of two images at particular location. In this proposed work, the difference between the pixel intensity at each location in **C** and **S** is taken and squared. Further for taking the mean square error this resultant value is divided by the total number of pixels present in an image.

$$MSE = \frac{1}{W \times H} \sum_{i=1}^{W} \sum_{i=1}^{H} \left(S\left(i,j\right) - C\left(i,j\right)\right)^2 \quad (3)$$

Here, $S\left(i,j\right) - C\left(i,j\right)$ is error and $W \times H$ is the one image size or total number of pixels in an image. Its value can range from 0 to $\infty$. For two similar images its value is 0 while for cancelable biometric based system its value should far from 0. Higher the distortion among two images higher will be the value of MSE.

- **Root Mean Square Error**: It is a standard way of representing the mean square error in quantitative data. It is square root of mean square error as depicted in below equation. Like MSE, the value of RMSE also ranges between 0 to $\infty$. The value of RMSE for two similar images is 0 while for cancelable biometric based system its value should far from 0. Higher value of RMSE denotes the two images are highly distorted to each other.

$$RMSE = \sqrt{MSE} \quad (4)$$

Here, MSE is Mean Square Error.

- **Peak Signal to Noise ratio**: It is the ratio between the maximum possible intensity value power of an image and the power of noise that effect the image quality. In other words, it is ratio between maximum pixel intensity value and Root Mean Square Error.

$$PSNR\left(db\right) = 20 \log_{10} \frac{255}{\sqrt{MSE}} \quad (5)$$

Here, 255 is the maximum value of single channel. The relation between PSNR and MSE can be seen in above relation Which shows that if MSE between **C** and **S** is high then PSNR value will be low and vice versa. In case of cancelable biometric system, the value of MSE should be high which results in lower value of PSNR i.e. noise among the two images should be high.

- **Structural Similarity**: It is another measure to represent how much difference is present between the two images. The value range of SSIM is between $-1$ to 1. For two similar images the SSIM value is 1. In case of cancelable biometric system the value of SSIM should approach to 0.

$$SSIM(S,C) = \frac{(2\mu_S\mu_C + T_1)(2\sigma_{SC} + T_2)}{(\mu_S^2 + \mu_C^2 + T_1)(\sigma_S^2 + \sigma_C^2 + T_2)} \quad (6)$$

Here $\mu_S$, $\mu_C$ are mean and $\sigma_S$, $\sigma_C$ are variance and $\sigma_{SC}$ is covariance between original image **S** and cancelable template **C**. However, $T_1$ and $T_2$ are constants

- **Mean Absolute Error**: It is an absolute average value used to represent the distortion provided in the original biometric image. Its value also ranges between 0 to $\infty$. For two similar images the value of MAE is 0 while for dissimilar images its value far from 0. In case of cancelable biometric system the value of MAE should be maximum.

$$MAE = \frac{1}{W \times H} \sum_{i=1}^{W} \sum_{i=1}^{H} |S(i,j) - C(i,j)| \quad (7)$$

Here, $S(i,j) - C(i,j)$ is error between original image and corresponding cancelable biometric template. While $W \times H$ is the image size or total number of pixels in an image

- **Number of Pixel Change Rate**: This measure is used to calculate the randomness between the two images. For two dissimilar images the value of NPCR is high. In cancelable biometric, its value should be high which represents the high extent of distortion in the cancelable template.

$$NPCR = (S_i, C_j) = \frac{\sum_{i,j} D(i,j)}{W \times H} \times 100\% \quad (8)$$

Here (i) $D(i,j) = 0$; if $I(i,j) = S(i,j)$ and (ii) $D(i,j) = 1$; if $I(i,j) \neq S(i,j)$.

- **Unified Average Changing Intensity**: It is normalized form of the mean absolute error. It is also used to check the difference between the intensity value among two images. In cancelable biometric system, the value of UACI should be high. From our experimental results, it has been observed that some state-of-the-art methods have high value of UACI but still they are unable to hide glimpses of original biometric in corresponding cancelable templates.

$$UACI = \frac{1}{W \times H} \times \frac{\sum_{i=1}^{W} \sum_{i=1}^{H} |I(i,j) - S(i,j)|}{255} \times 100\% \quad (9)$$

Here, 255 is maximum intensity value in single channel and $I(i,j) - S(i,j)$ is error while $W \times H$ is image size

Table 3: Quantitative results for gray images on CP ear, UTIRIS iris, ORL face and IIT Delhi iris datasets using proposed Algorithm 2

| Methods | Cr | MAE | NPCR | PSNR | RMSE | SSIM | UACI |
|---|---|---|---|---|---|---|---|
| **CP ear** | | | | | | | |
| Bin Salt | 0.3685 | 26.6611 | 35.4147 | 29.7650 | 9.0401 | 0.3141 | 0.03510 |
| Gray Salt | 0.2874 | 0.1326 | 38.4164 | Inf | 0.4737 | 0.3019 | 0.00003 |
| RBit-XOR | 0.3178 | 18.1543 | 51.3793 | 29.4138 | 8.6877 | 0.1744 | 0.02750 |
| Index of Max Hashing | -0.0033 | 41.4980 | 49.9003 | 30.5797 | 8.4266 | 0.0247 | 24.11690 |
| Fused Structure | 0.1568 | 38.7793 | 50.0025 | 30.9976 | 8.8739 | 0.0283 | 23.05070 |
| Permutation Indexing | 0.0005 | 31.6737 | 99.5364 | 29.6212 | 7.9331 | 0.0295 | 16.34260 |
| Proposed | -0.0012 | 31.4192 | 99.9895 | 27.8996 | 10.3203 | 0.0103 | 12.32120 |
| **UTIRIS** | | | | | | | |
| Bin Salt | 0.3358 | 15.4143 | 42.4160 | 31.4486 | 7.3453 | 0.3016 | 0.02570 |
| Gray Salt | 0.2707 | 0.0055 | 35.4165 | 68.6364 | 0.1569 | 0.2247 | 0.00001 |
| RBit-XOR | 0.3359 | 21.5205 | 51.3869 | 28.6801 | 9.3937 | 0.1845 | 0.03770 |
| Index of Max Hashing | -0.0148 | 60.6443 | 49.9675 | 28.2759 | 9.8967 | 0.0262 | 27.70360 |
| Fused Structure | -0.1283 | 55.4433 | 50.0502 | 27.7256 | 9.9323 | 0.0230 | 21.74250 |
| Permutation Indexing | 0.0015 | 29.8329 | 99.3266 | 28.4457 | 9.8249 | 0.0302 | 11.69920 |
| Proposed | -0.0023 | 36.3812 | 99.9956 | 27.5930 | 10.6848 | 0.0101 | 14.26720 |
| **ORL face** | | | | | | | |
| Bin Salt | 0.3575 | 18.2217 | 36.4149 | 29.8268 | 8.4661 | 0.3436 | 0.03160 |
| Gray Salt | 0.3607 | 0.0270 | 24.4163 | 57.4669 | 0.4177 | 0.2505 | 0.00009 |
| RBit-XOR | 0.3504 | 25.9038 | 47.3863 | 27.8707 | 10.3721 | 0.1970 | 0.05080 |
| Index of Max Hashing | -0.0055 | 54.0264 | 49.9366 | 28.5310 | 9.4937 | 0.0355 | 25.10840 |
| Fused Structure | 0.1574 | 50.0738 | 49.9951 | 28.9516 | 9.9148 | 0.0263 | 23.55830 |
| Permutation Indexing | -0.0017 | 23.7709 | 99.4188 | 28.3948 | 9.7281 | 0.0292 | 9.32190 |
| Proposed | -0.0025 | 29.3690 | 99.9956 | 28.1091 | 10.0471 | 0.0098 | 11.51720 |
| **IITD iris** | | | | | | | |
| Bin Salt | 0.3275 | 18.3122 | 49.5682 | 32.0333 | 6.8382 | 0.3043 | 7.18130 |
| Gray Salt | 0.2879 | 0.0005 | 46.1307 | Inf | 0.0009 | 0.1846 | 0.00080 |
| RBit-XOR | 0.3935 | 25.0757 | 43.2950 | 28.4284 | 9.6652 | 0.2394 | 9.83360 |
| Index of Max Hashing | -0.0059 | 95.3870 | 48.0685 | 27.1983 | 14.0205 | 0.0075 | 37.40670 |
| Fused Structure | -0.0577 | 40.5832 | 49.8713 | 30.3754 | 9.0572 | 0.0290 | 15.91500 |
| Permutation Indexing | 0.0061 | 36.5739 | 98.8301 | 27.5929 | 10.6394 | 0.0143 | 14.34270 |
| Proposed | -0.0031 | 42.1365 | 99.9898 | 26.3531 | 10.9573 | 0.0070 | 16.52410 |

The quantitative results of above performance measures obtained from proposed methods and state-of-the-art methods are given in Tables 9 and 10 for gray datasets while Tables 5 and 6 represents results for color datasets. It can be noticed from Tables 9 and 10 that the average Correlation between the original biometric and cancelable biometric template, generated by proposed methods, is almost zero which signifies that these two entities are not correlated. Although, for other compared methods the correlation between **S** and **C** is present. Further, the value of NPCR is highest for the proposed methods while lowest for PSNR. The RMSE is also maximum for the proposed methods while SSIM value is minimum which signifies that two patterns are structurally dissimilar. Hence, it can be seen that proposed methods outperform all the compared methods in terms of almost all performance measures except one or two on a variety of biometrics. Similar observations can be made on the experimental results on color images as shown in Tables 5 and 6 respectively. The red and blue entries in a column denote the method stood at first and second place respectively in terms of various performance measures. Both proposed methods have been compared with six other methods i.e. Bin Salt [27], Gray Salt [27], RBit-XOR [20], Index of Max Hashing [9], Fused Structure [10] and Permutation Indexing [7]. In a single result table, total 28 first and 28 second comparisons are present. This shows number of times a single method stood first and second places among total of 28 first and 28 second places. Total four Tables are present for all datasets (both gray and color). In this way, a total of 112 (= 28 first/second places in single Table × 4 Tables) first places and 112 second places are available in all the tables. Number of times individual method stood _rst and second place in all Tables is depicted by bar graph as shown in Fig. 9. From the bar graph, it can be seen that the proposed methods stood at first place 55 times out of 112 first places and second place 23 times out of 112 second places. Similarly, Index of Max Hashing stood 32 times first and 30 times second, Permutation Indexing stood 13 times first and 19 times second, Fused Structure stood 8 times first and 36 times second. However, RBit-XOR stood four times first and two times second, Bin Salt stood two times second while Gray Salt didn't secure a single place in terms of performance measures.

Table 4: Quantitative results for gray images on CP ear, UTIRIS iris, ORL face and IIT Delhi iris datasets using proposed Algorithm 3

| Methods | Cr | MAE | NPCR | PSNR | RMSE | SSIM | UACI |
|---|---|---|---|---|---|---|---|
| | | | CP ear | | | | |
| Bin Salt | 0.3685 | 26.6611 | 35.4147 | 29.7650 | 9.0401 | 0.3141 | 0.03510 |
| Gray Salt | 0.2874 | 0.1326 | 38.4164 | Inf | 0.4737 | 0.3019 | 0.00003 |
| RBit-XOR | 0.3178 | 18.1543 | 51.3793 | 29.4138 | 8.6877 | 0.1744 | 0.02750 |
| Index of Max Hashing | -0.0033 | 41.4980 | 49.9003 | 30.5797 | 8.4266 | 0.0247 | 24.11690 |
| Fused Structure | 0.1568 | 38.7793 | 50.0025 | 30.9976 | 8.8739 | 0.0283 | 23.05070 |
| Permutation Indexing | 0.0005 | 31.6737 | 99.5364 | 26.6212 | 7.9331 | 0.0295 | 16.34260 |
| Proposed | 0.0023 | 31.2946 | 99.6106 | 27.9150 | 10.3016 | 0.0104 | 12.27290 |
| | | | UTIRIS | | | | |
| Bin Salt | 0.3358 | 15.4143 | 42.4160 | 31.4486 | 7.3453 | 0.3016 | 0.02570 |
| Gray Salt | 0.2707 | 0.0055 | 35.4165 | 68.6364 | 0.1569 | 0.2247 | 0.00001 |
| RBit-XOR | 0.3359 | 21.5205 | 51.3869 | 28.6801 | 9.3937 | 0.1845 | 0.03770 |
| Index of Max Hashing | -0.0148 | 60.6443 | 49.9675 | 28.2759 | 9.8967 | 0.0262 | 27.70360 |
| Fused Structure | -0.1283 | 55.4433 | 50.0502 | 27.7256 | 9.9323 | 0.0230 | 21.74250 |
| Permutation Indexing | 0.0015 | 29.8329 | 99.3266 | 28.4457 | 9.8249 | 0.0302 | 11.69920 |
| Proposed | 0.0028 | 36.2451 | 99.6096 | 27.6059 | 10.6688 | 0.0102 | 14.21370 |
| | | | ORL face | | | | |
| Bin Salt | 0.3575 | 18.2217 | 36.4149 | 29.8268 | 8.4661 | 0.3436 | 0.03160 |
| Gray Salt | 0.3607 | 0.0270 | 24.4163 | 57.4669 | 0.4177 | 0.2505 | 0.00009 |
| RBit-XOR | 0.3504 | 25.9038 | 47.3863 | 27.8707 | 10.3721 | 0.1970 | 0.05080 |
| Index of Max Hashing | -0.0055 | 54.0264 | 49.9366 | 28.5310 | 9.4937 | 0.0355 | 25.10840 |
| Fused Structure | 0.1574 | 50.0738 | 49.9951 | 28.9516 | 9.9148 | 0.0263 | 23.55830 |
| Permutation Indexing | -0.0017 | 23.7709 | 99.4188 | 28.3948 | 9.7281 | 0.0292 | 9.32190 |
| Proposed | 0.0030 | 29.2166 | 99.6133 | 28.1275 | 10.0250 | 0.0100 | 11.45750 |
| | | | IITD iris | | | | |
| Bin Salt | 0.3275 | 18.3122 | 49.5682 | 32.0333 | 6.8382 | 0.3043 | 7.18130 |
| Gray Salt | 0.2879 | 0.0005 | 46.1307 | Inf | 0.0009 | 0.1846 | 0.00080 |
| RBit-XOR | 0.3935 | 25.0757 | 43.2950 | 28.4284 | 9.6652 | 0.2394 | 9.83360 |
| Index of Max Hashing | -0.0059 | 95.3870 | 48.0685 | 27.1983 | 14.0205 | 0.0075 | 37.40670 |
| Fused Structure | -0.0577 | 40.5832 | 49.8713 | 30.3754 | 9.0572 | 0.0290 | 15.91500 |
| Permutation Indexing | 0.0061 | 36.5739 | 98.8301 | 27.5929 | 10.6394 | 0.0143 | 14.34270 |
| Proposed | 0.0032 | 41.9214 | 99.6133 | 26.3722 | 10.9329 | 0.0065 | 16.43970 |

Table 5: Quantitative results for color images on CP ear, UTIRIS iris, AR face and IIT Delhi iris datasets using proposed Algorithm 2

| Methods | Cr | MAE | NPCR | PSNR | RMSE | SSIM | UACI |
|---|---|---|---|---|---|---|---|
| **CP ear** | | | | | | | |
| BIN Salt | 0.3726 | 16.1243 | 60.3736 | Inf | 8.7680 | 0.2136 | 21.6780 |
| Gray Salt | 0.3419 | 17.3465 | 96.9898 | Inf | 8.6750 | 0.6716 | 20.5436 |
| RBit-XOR | 0.4196 | 18.7189 | 98.4661 | 28.7543 | 9.3166 | 0.0552 | 22.0222 |
| Index of Max Hashing | 0.2192 | 91.6883 | 72.4308 | 27.0778 | 11.2889 | 0.0105 | 80.8686 |
| Fused Structure | -0.0626 | 75.7431 | 98.4153 | 27.9853 | 10.1690 | 0.0060 | 81.1095 |
| Permutation Indexing | 0.0002 | 34.3336 | 98.7996 | 27.3449 | 11.0118 | 0.0286 | 40.3925 |
| Proposed | 0.0024 | 70.2292 | 99.7851 | 25.7906 | 13.1121 | 0.0102 | 82.6226 |
| **UTIRIS** | | | | | | | |
| Bin Salt | 0.2892 | 11.5604 | 97.8171 | 32.9566 | 6.3329 | 0.2832 | 13.6005 |
| Gray Salt | 0.3683 | 0.0205 | 97.1941 | 58.5614 | 0.3726 | 0.2679 | 0.0241 |
| RBit-XOR | 0.4304 | 20.9749 | 98.5574 | 28.0217 | 10.1323 | 0.2249 | 24.6764 |
| Index of Max Hashing | 0.0007 | 73.4919 | 99.3238 | 28.1113 | 9.1623 | 0.0247 | 86.4611 |
| Fused Structure | 0.2965 | 58.3934 | 99.0590 | 28.9269 | 8.0080 | 0.2867 | 68.6981 |
| Permutation Indexing | 0.0009 | 31.7332 | 99.1744 | 27.3341 | 10.9630 | 0.0321 | 37.3332 |
| Proposed | -0.0024 | 31.6639 | 100.0000 | 27.0757 | 11.0936 | 0.0096 | 37.2516 |
| **AR face** | | | | | | | |
| Bin Salt | 0.2867 | 17.6261 | 95.7843 | 29.6312 | 8.6082 | 0.2683 | 20.7366 |
| Gray Salt | 0.3151 | 0.0084 | 95.2613 | 68.2847 | 0.1932 | 0.2331 | 0.0099 |
| RBit-XOR | 0.3603 | 18.7915 | 92.1341 | 29.0458 | 9.0152 | 0.2134 | 22.1076 |
| Index of Max Hashing | 0.0009 | 59.6217 | 99.6712 | 29.3985 | 9.7028 | 0.0264 | 70.1432 |
| Fused Structure | 0.3814 | 64.0564 | 96.0148 | 29.8593 | 9.1332 | 0.2562 | 75.3605 |
| Permutation Indexing | -0.0003 | 31.1703 | 99.0873 | 27.1549 | 11.1931 | 0.0358 | 36.6710 |
| Proposed | -0.0040 | 40.5696 | 99.8917 | 27.1298 | 11.2967 | 0.0186 | 47.7289 |
| **IITD iris** | | | | | | | |
| Bin Salt | 0.3281 | 18.1737 | 98.6803 | 32.0936 | 6.8063 | 0.3042 | 21.3808 |
| Gray Salt | 0.2856 | 0.0765 | 91.9948 | Inf | 0.0356 | 0.1844 | 0.0897 |
| RBit-XOR | 0.3894 | 25.1371 | 86.2976 | 28.4296 | 9.6640 | 0.2389 | 29.5731 |
| Index of Max Hashing | 0.0011 | 77.0563 | 99.0258 | 27.4080 | 13.6843 | 0.0111 | 90.6545 |
| Fused Structure | 0.3506 | 63.6589 | 99.3264 | 26.1429 | 12.7390 | 0.3004 | 50.8928 |
| Permutation Indexing | 0.0009 | 36.7071 | 98.8589 | 27.5912 | 10.6413 | 0.0144 | 43.1848 |
| Proposed | -0.0030 | 42.1678 | 100.0000 | 27.3507 | 10.9602 | 0.0086 | 49.6092 |

## 4.4 How Performance Measures Satisfies the Characteristics of Cancelable Biometric?

The main characteristics of cancelable biometric system is that the generated template should be distorted version of the original biometric. It should not reveal any information regarding its original credentials. In this proposed work, seven performance measures have been used for finding the relationship between original image (**S**) and cancelable biometric template (**C**) generated corresponding to the original image. The relation between performance measures and cancelable biometric characteristics is discussed below which is inspired from research work of Trivedi et. al [30]:

Table 6: Quantitative results for color images on CP ear, UTIRIS iris, AR face and IIT Delhi iris datasets by proposed Algorithm 3

| Methods | Cr | MAE | NPCR | PSNR | RMSE | SSIM | UACI |
|---|---|---|---|---|---|---|---|
| **CP ear** | | | | | | | |
| BIN Salt | 0.3726 | 16.1243 | 60.3736 | Inf | 8.7680 | 0.2136 | 21.6780 |
| Gray Salt | 0.3419 | 17.3465 | 96.9898 | Inf | 8.6750 | 0.6716 | 20.5436 |
| RBit-XOR | 0.4196 | 18.7189 | 98.4661 | 28.7543 | 9.3166 | 0.0552 | 22.0222 |
| Index of Max Hashing | 0.2192 | 91.6883 | 72.4308 | 27.0778 | 11.2889 | 0.0105 | 80.8686 |
| Fused Structure | -0.0626 | 75.7431 | 98.4153 | 27.9853 | 10.1690 | 0.0060 | 81.1095 |
| Permutation Indexing | 0.0002 | 34.3336 | 98.7996 | 27.3449 | 11.0118 | 0.0286 | 40.3925 |
| Proposed | 0.0022 | 70.2373 | 99.8110 | 25.7899 | 13.1132 | 0.0103 | 82.6322 |
| **UTIRIS** | | | | | | | |
| Bin Salt | 0.2892 | 11.5604 | 97.8171 | 32.9566 | 6.3329 | 0.2832 | 13.6005 |
| Gray Salt | 0.3683 | 0.0205 | 97.1941 | 58.5614 | 0.3726 | 0.2679 | 0.0241 |
| RBit-XOR | 0.4304 | 20.9749 | 98.5574 | 28.0217 | 10.1323 | 0.2249 | 24.6764 |
| Index of Max Hashing | 0.0007 | 73.4919 | 99.3238 | 28.1113 | 9.1623 | 0.0247 | 86.4611 |
| Fused Structure | 0.2965 | 58.3934 | 99.0590 | 28.9269 | 8.0080 | 0.2867 | 68.6981 |
| Permutation Indexing | 0.0009 | 31.7332 | 99.1744 | 27.3341 | 10.9630 | 0.0321 | 37.3332 |
| Proposed | 0.0020 | 31.5300 | 99.6074 | 27.0897 | 10.9972 | 0.0098 | 37.0941 |
| **AR face** | | | | | | | |
| Bin Salt | 0.2867 | 17.6261 | 95.7843 | 29.6312 | 8.6082 | 0.2683 | 20.7366 |
| Gray Salt | 0.3151 | 0.0084 | 95.2613 | 68.2847 | 0.1932 | 0.2331 | 0.0099 |
| RBit-XOR | 0.3603 | 18.7915 | 92.1341 | 29.0458 | 9.0152 | 0.2134 | 22.1076 |
| Index of Max Hashing | 0.0009 | 59.6217 | 99.6712 | 29.3985 | 9.7028 | 0.0264 | 70.1432 |
| Fused Structure | 0.3814 | 64.0564 | 96.0148 | 29.8593 | 9.1332 | 0.2562 | 75.3605 |
| Permutation Indexing | -0.0003 | 31.1703 | 99.0873 | 27.1549 | 11.1931 | 0.0358 | 36.6710 |
| Proposed | 0.0038 | 40.3595 | 99.8984 | 27.1335 | 11.2915 | 0.0189 | 47.4818 |
| **IITD iris** | | | | | | | |
| Bin Salt | 0.3281 | 18.1737 | 98.6803 | 32.0936 | 6.8063 | 0.3042 | 21.3808 |
| Gray Salt | 0.2856 | 0.0765 | 91.9948 | Inf | 0.0356 | 0.1844 | 0.0897 |
| RBit-XOR | 0.3894 | 25.1371 | 86.2976 | 28.4296 | 9.6640 | 0.2389 | 29.5731 |
| Index of Max Hashing | 0.0011 | 77.0563 | 99.0258 | 27.4080 | 13.6843 | 0.0111 | 90.6545 |
| Fused Structure | 0.3506 | 63.6589 | 99.3264 | 26.1429 | 12.7390 | 0.3004 | 50.8928 |
| Permutation Indexing | 0.0009 | 36.7071 | 98.8589 | 27.5912 | 10.6413 | 0.0144 | 43.1848 |
| Proposed | 0.0034 | 41.9519 | 99.6082 | 27.3683 | 10.9379 | 0.0093 | 49.3551 |

- ***Diversity*** property in cancelable biometric template states that templates generated corresponding to single identity should be different. In this proposed work, cancelable templates are generated using BitXOR between original biometric $\mathbf{S}$ and random walk matrix $\mathbf{R}_W$. Further, this $\mathbf{R}_W$ matrix is obtained from random matrix $\mathbf{R}$. Hence for generating the different cancelable templates for same user, this random matrix needs to change which result in different templates for same user. The difference between old and new templates can be found by above discussed seven performance measures. The diversity between two templates generated for same user is shown in Table 7. From this Table, it can be seen that results are satisfying the dissimilar properties of two images. Hence, two templates generated for same user are different to each other. The results are presented for cancelable biometric templates generated by proposed Algorithm 2 for gray datasets. Similar results have been obtained for templates generated with Algorithm 3 and color datasets.

- **Non-invertible** property states that intruder should not able to retrieve the original credentials after accessing the cancelable templates. It simply means no reverse engineering process could reveal the original credential from the distorted ones. For satisfying this property, we have used computational complexity method which shows how much computational difficult is to obtain the original credential from cancelable template generated by proposed method. For generating the cancelable biometric templates three important steps are there. (i) Uniform distributed random image **R** (ii) Random walk matrix $\mathbf{R}_W$ and (iii) Cancelable template generation using BitXOR operation. In this proposed work, 320×240 uniform distributed random image is used, from which by using proposed Algorithms another same size $\mathbf{R}_W$ matrix is generated. Finally, this $\mathbf{R}_W$ matrix is operated with original biometric **S** for generation of cancelable biometric template. Hence, the proposed method consists with many steps for providing distortion in original biometric. A rough computation complexity can be seen as 320×240 is the image size which results in total number of 76800 pixels and each image is represented by 8 bits per pixel. So the possible combination will be $76800^{256}$. This results in very huge number of images. If the number of bits required to represent single pixel is 16 bits then complexity will be $76800^{65536}$. In case of color image, these combinations will become $76800^{16777216}$ (8 bits for single pixel i in each channel so total channels are three) i.e. RGB (3×8=24 bits for each pixel). This is for one image e.g. random image. For random walk matrix, same complexity will be applicable which results in $76800^{256}$ and $76800^{16777216}$ combinations for gray and color image respectively. Hence, the total complexity will be $2 \times 76800^{256}$ and $2\_76800^{16777216}$ for gray and color image. This means these number of combination will be required for getting the glimpse of the original biometric which is not computationally feasible.

-**Revocable** property states that in case of adversary attack the previous allotted cancelable template should be discarded and a new should be allotted to the genuine user. In this proposed work, random walk matrix $\mathbf{R}_W$ is used to provide the distortion in the original biometric which in turn depends on the random matrix **R** generated using uniform distribution method. With the help of this $\mathbf{R}_W$ matrix, we can generate many cancelable biometric template and given to the genuine user in case of compromise. For experimental work, we have used 320×240 size images. For gray image, we can generate $76800^{256}$ (for 8 bits per pixel representation) number of possible combinations for uniform distributed random image **R**. For color image, these combinations will become $76800^{16777216}$ (24 bits for each pixel), which are very huge in count. By selecting any one of these combination (except the one used before), different cancelable biometric template can be allotted to the genuine user in case of compromise or adversarial attack.

- **Performance**: The cancelable biometric templates generated by proposed methods are secure than templates generated by state-of-the-art methods as proved by quantitative and qualitative analysis. Hence it can be ensured that performance of the system will not deteriorate.

Table 7: Performance measures used for showing the diversity property between two cancelable template generated for same user. The two templates are generated for same user using different random matrix **R**

| Datasets | Cr | MAE | NPCR | PSNR | RMSE | SSIM | UACI |
|---|---|---|---|---|---|---|---|
| CP ear | 0.0031 | 42.5028 | 99.6484 | 27.4472 | 10.8189 | 0.0090 | 16.6678 |
| UTIRIS | 0.0012 | 42.6492 | 99.6068 | 27.4517 | 10.8133 | 0.0069 | 16.7252 |
| ORL face | 0.0011 | 42.5290 | 99.6185 | 27.4431 | 10.8239 | 0.0070 | 16.6781 |
| IITD iris | -0.0001 | 42.9074 | 99.6328 | 27.4377 | 10.8306 | 0.0054 | 16.8264 |

## 4.5 Discussion

Here, we critically analyze the experimental results presented in Tables 9-6. As already described, the proposed methods work equally well on gray as well as color biometric images. The proposed methods outperform all other methods in qualitative as well as quantitative terms. A qualitative analysis was done in subsection 4.2 where the original biometric and the corresponding cancelable templates were shown. The reason behind other state-of-the-art methods revealing partial and full information about original biometric is that these methods use the original biometric along with other natural images while the proposed methods introduce randomness in the original image using random walk model. Further, histogram analysis shows that cancelable templates generated by proposed methods do not reveal any information regarding original image. This strengthens the fact that for any intruder, it would be extremely difficult to know about the original image. Afterwards, a quantitative analysis has also been presented for both gray and color datasets. A comparison between cancelable biometric templates generated by proposed methods and other methods has been presented based on several performance measures including Cr, RMSE, PSNR, SSIM, MAE, NPCR and UACI. Bar graph as shown in Fig. 9 represents which method performed better than others in terms of quantitative performance measures. In the bar graph, x-axis shows the method name while y-axis denotes the number of times a method stood first or second in quantitative results. It is clear from the bar graph that proposed methods stood first at many places.

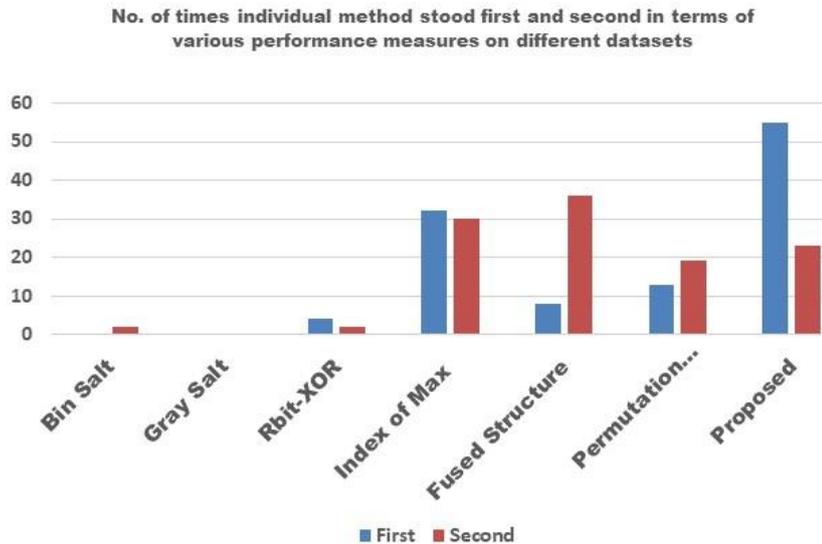

Fig. 9: Number of times following methods i.e. Bin Salt [27], Gray Salt [27], RBit-XOR [20], Index of Max Hashing [9], Fused Structure [10], Permutation Indexing [7] and Proposed method stood at _rst and second place in terms of quantitative measures

## 4.6 Comparison of proposed method with state-of-the-art methods

In Bin Salt [27] and Gray Salt [27] methods, auxiliary data is added with original biometric to generate a cancelable template and this addition of auxiliary data is known as salting method. The main advantage of this method is simple for implementation while main disadvantage is to know the amount of auxiliary data needed to add with the original biometric image for cancellable template generation. Proposed methods are different from other methods in terms of security as template generated by these methods reveal partial and full information about the original biometric as depicted by Figs 5-6. Both Bin Salt and Gray Salt methods used simple addition and subtraction operations which are invertible in nature. Both these methods are tested on only gray iris dataset. Proposed methods are different from the Index of Max Hashing [9] and Fused Structure [10] methods on the basis of quality of templates, limited number of biometric modalities and feature sets. For different datasets, templates generated by these methods reveal partial and full information about original credentials which is not reveal by proposed method. A comparative study among all methods is presented in Table 8. Further, comparison among proposed method with the state-of-the-art methods is also carried out in terms of histogram analysis.

Table 8: Comparison of the Proposed method with state-of-the-art methods

| Factor | Bin Salt | Gray Salt | RBit-XOR | Index of Max | Fused Structure | Permutation Indexing | Proposed |
|---|---|---|---|---|---|---|---|
| Random Image | | | | | | | √ |
| Random Walk | | | | | | | √ |
| BitXOR | | | √ | | | | √ |
| Gray Images | √ | √ | √ | √ | √ | √ | √ |
| Color Images | | | | | | | √ |
| Histogram Analysis | | | | | | | √ |
| Number of Datasets | 1 | 1 | 1 | 2 | 2 | 3 | 8 |
| Biometric Modalities | iris | iris | iris | fingerprint | fingerprint | face, fingerprint, fingervein | ear, face, iris |
| Glimpses in templates | √ | √ | √ | √ | √ | √ | |
| Limited feature sets | √ | √ | | √ | √ | √ | |

## 5 Conclusion and Future Work

This research work has proposed two novel methods for generation of cancelable biometric templates using random walk method. Towards this effect two methods are proposed viz. (a) Cancelable Biometric template generation based on Random Walk using Bit XOR (CBRW-

BitXOR) (b) Cancelable Biometric template generation based on Random Walk using Bit Complement (CBRW-BitCMP) have been suggested and two novel algorithms have been proposed. Both of these algorithms employ another proposed algorithm for generation of random walk matrix. Extensive experiments are performed on a various biometric datasets including ear, iris and face. Proposed methods have been compared with other methods by qualitative and quantitative analysis. The qualitative analysis shows that proposed methods are able to transform original credentials without revealing them while other compared methods do reveal some structure of the original biometric. The quantitative analysis shows that the proposed methods outperform other compared methods in terms of seven performance measures i.e. Cr, MAE, RMSE, PSNR, UACI, SSIM and NPCR. The advantages of the proposed methods include: (i) Independence from the fact whether original biometric image is gray or color and (ii) No information about original credentials is revealed by proposed methods. In future work, new methods will be explored which result better in both qualitative and quantitative measures.